\documentclass[11pt]{article}  % <===== This line must be replaced for conference papers

\usepackage[final]{automl}
\usepackage{microtype} % microtypography
\usepackage{booktabs}  % tables
\usepackage{url}  % urls
\usepackage{csquotes}
\usepackage{natbib}

\usepackage[utf8]{inputenc}
\usepackage{url}

\usepackage{amsmath}
\usepackage{bbm}
\usepackage{bm}
\usepackage{cancel}
\usepackage{mathtools}

\usepackage{ascmac}  % ascmac must come before `algorithm' and `algpseudocode'
\usepackage{algorithm}
\usepackage{algpseudocode}

\usepackage{fancyhdr}
\usepackage{tcolorbox}
\usepackage{tikz}

\usepackage{booktabs}
\usepackage{comment}
\usepackage{lastpage}
\usepackage{listings}
\usepackage{multibib}
\usepackage{multirow}
\usepackage[super]{nth}

\tcbuselibrary{breakable, skins, theorems}

\allowdisplaybreaks

\newtheorem{proposition}{Proposition}

\newtheorem{proof}{Proof}

\algtext*{EndFor}
\algtext*{EndWhile}
\algtext*{EndIf}
\algtext*{EndProcedure}
\algtext*{EndFunction}
\algnewcommand{\LineComment}[1]{\State \(\triangleright\) #1}

\newcites{appx}{References}

\hypersetup{%
  pdfauthor={Shuhei Watanabe}, % will be reset to "Anonymous" unless the "final" package option is given
  pdftitle={Box Decomposition Algorithms Explained},
  pdfsubject={},
  pdfkeywords={AutoML, LaTeX, style}
}

\newif\ifunderreview
\newif\ifsubmission
\newif\ifappendix

\underreviewfalse

\submissiontrue
\appendixfalse

\ifsubmission
\newcommand{\todo}[1]{}
\newcommand{\replace}[2]{}
\newcommand{\sw}[1]{}

\else
\newcommand{\todo}[1]{\textbf{\textcolor{red}{[TODO: #1]}}}

\newcommand{\replace}[2]{\textbf{\textcolor{red}{[del: \cancel{#1}]}}\textbf{\textcolor{blue}{[new: #2]}}}

\newcommand{\sw}[1]{\textbf{\textcolor{cyan}{[SW: #1]}}}

\usepackage[inline]{showlabels}

\fi

\makeatletter
\newcommand{\customlabel}[2]{%
   \protected@write \@auxout {}{\string \newlabel {#1}{{#2}{\thepage}{#2}{#1}{}} }%
   \hypertarget{#1}{}
}
\makeatother

\newcommand{\argmax}{\mathop{\mathrm{argmax}}}

\newcommand{\floor}[1]{\lfloor #1 \rfloor}

\renewcommand{\eqref}[1]{Eq.~(\ref{#1})}

\DeclareFixedFont{\ttb}{T1}{txtt}{bx}{n}{12} % for bold
\DeclareFixedFont{\ttm}{T1}{txtt}{m}{n}{12}  % for normal

\definecolor{deepblue}{rgb}{0,0,0.5}
\definecolor{deepred}{rgb}{0.6,0,0}
\definecolor{deepgreen}{rgb}{0,0.5,0}

\newcommand{\yv}{\boldsymbol{y}}
\newcommand{\rv}{\boldsymbol{r}}
\newcommand{\Hcal}{\mathcal{H}}

\begin{document}
\title{Approximation of Box Decomposition Algorithm for Fast Hypervolume-Based Multi-Objective Optimization}

\author{
  Shuhei Watanabe \\
  Preferred Networks Inc. \\
  \texttt{shuheiwatanabe@preferred.jp}
}

\maketitle

\begin{abstract}
  Hypervolume (HV)-based Bayesian optimization (BO) is one of the standard approaches for multi-objective decision-making.
  However, the computational cost of optimizing the acquisition function remains a significant bottleneck, primarily due to the expense of HV improvement calculations.
  While HV box-decomposition offers an efficient way to cope with the frequent exact improvement calculations, it suffers from super-polynomial memory complexity $O(MN^{\floor{\frac{M + 1}{2}}})$ in the worst case as proposed by \cite{lacour2017box}.
  To tackle this problem, \cite{couckuyt2012towards} employed an approximation algorithm.
  However, a rigorous algorithmic description is currently absent from the literature.
  This paper bridges this gap by providing comprehensive mathematical and algorithmic details of this approximation algorithm.
\end{abstract}

\begin{figure}[t]
  \centering
  \includegraphics[width=0.98\textwidth]{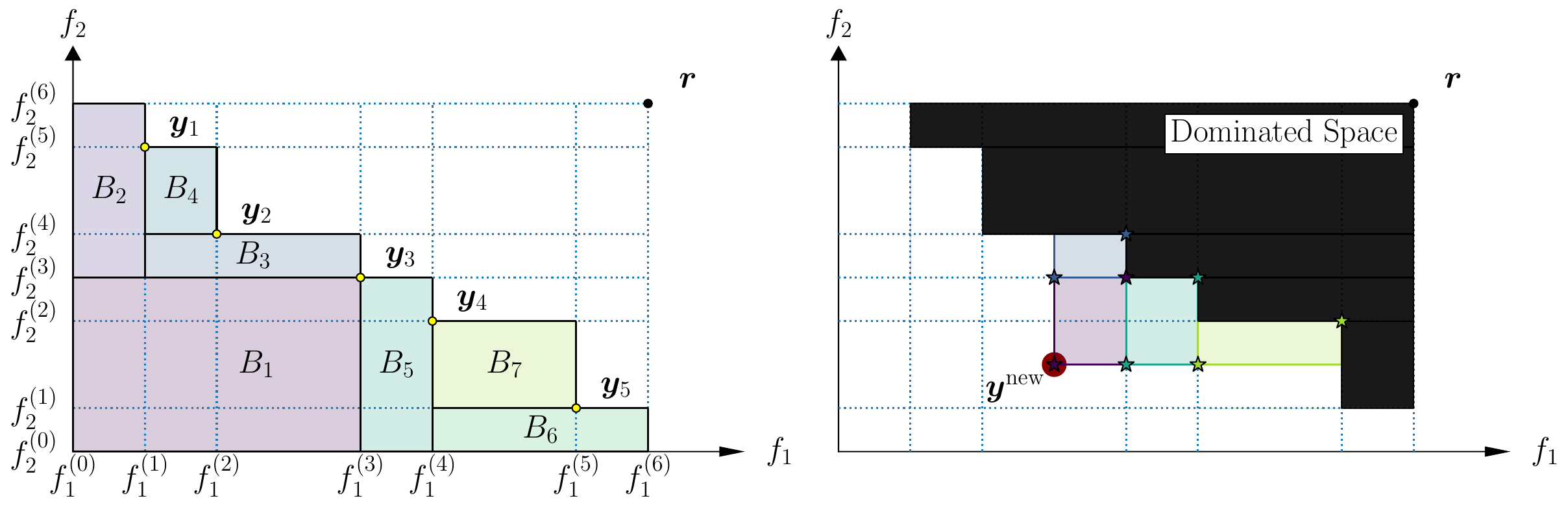}
  \caption{
    Conceptual visualizations for the notations used in this paper.
    \textbf{Left}: Visualization of box decomposition. The dominated space $\mathcal{D}$ is white, and the non-dominated space $\mathcal{N}$ is divided by hyperrectangles $B_k$.
    Box decomposition algorithms identify the lower and upper bounds of each hyperrectangle.
    \textbf{Right}: Visualization of HV improvement calculation.
    The HV improvement calculation is the summation of the HV for each intersection of hyperrectangle $B_k$ and the new entry $\yv^{\mathrm{new}}$ once HBDA is performed.
    The intersections with $B_1,B_3,B_5,B_7$ exhibit non-zero HVs in this example.
  }
  \label{fig:box-decomp-conceptual}
\end{figure}

\section{Preliminaries}

We first note that $\boldsymbol{a} \coloneqq [a_1, a_2, \dots, a_M] < \boldsymbol{b}\coloneqq [b_1, b_2, \dots, b_M]$ means $a_m < b_m$ for $m \in [M] \coloneqq \{1,\dots,M\}$.
Furthermore, we follow the dominance operator defined by \cite{watanabe2023python}:
\begin{itemize}
  \item $\yv$ dominates $\yv^\prime$, i.e., $\yv \prec \yv^\prime$, iff $\yv \leq \yv^\prime$ and $\yv \neq \yv^\prime$, and
  \item $Y = \{\yv_n\}_{n=1}^N$ dominates $\yv^\prime$ iff $\yv_n \prec \yv^\prime$ for $n \in [N]$.
\end{itemize}
We then introduce the notations used in this paper:
\begin{enumerate}
  \item $\rv \in \mathbb{R}^{M}$, a reference point where $M \in \mathbb{Z}_+$ is the number of objectives,
  \item $\mathcal{Y} \coloneqq \{\yv \in \mathbb{R}^M \mid \yv \preceq \rv\}$, the objective space, 
  \item $\mathcal{P} \coloneqq \{\yv_n\}_{n=1}^N \subseteq \mathcal{Y}$, a Pareto set where $N \in \mathbb{Z}_+$ is the number of Pareto solutions,
  \item $f_m^{(n)} \in \mathcal{Y}$, the $n$-th smallest value in the $m$-th objective among the Pareto set $\mathcal{P}$,
  \item $\mathcal{D} \coloneqq \{\yv \in \mathcal{Y} \mid \mathcal{P} \prec \yv \}$, the dominated space,
  \item $\mathcal{N} \coloneqq \mathcal{Y} \setminus \mathcal{D}$, the non-dominated space,
  \item $B_{k} \coloneqq \prod_{m=1}^M [l_{k,m}, u_{k,m}]\subseteq \mathcal{N}$, the $k$-th hyperrectangle defined on the non-dominated space where $B_i \cap B_j = \emptyset$ for $i \neq j$, $\bigcup_{k=1}^K B_k = \mathcal{N}$, and $K$ is the number of hyperrectangles, and
  \item $\Hcal(Y, \rv): \mathcal{B}_{y} \times \mathbb{R}^M \rightarrow \mathbb{R}_{\geq}$, the function that calculates the hypervolume of the space dominated $Y$, i.e., $\{\yv \in \mathcal{Y} \mid Y \prec \yv\}$, where $\mathcal{B}_y$ is the Borel set on $\mathcal{Y}$.
\end{enumerate}
We also define $f_m^{(0)} \coloneqq f_m^{(1)} - 1$ and $f_m^{(N+1)} \coloneqq f_m^{(N)} + 1$.
Figure~\ref{fig:box-decomp-conceptual} visualizes the notations intuitively.
The goal of HV box-decomposition algorithm (HBDA) is to compute all the hyperrectangles as fast as possible.

\section{Box-Decomposition Algorithm for Fast Hypervolume Improvement Computation}

Hypervolume (HV)-based Bayesian optimization (BO) methods, such as those proposed by \cite{daulton2020differentiable}, require frequent evaluations of the HV improvement $\Hcal(\mathcal{P} \cup \{\yv^{\mathrm{new}}\}, \rv) - \Hcal(\mathcal{P}, \rv)$.
When using HBDA, this improvement is calculated as $\sum_{k=1}^K \prod_{m=1}^{M} [u_{k,m} - \max(l_{k,m}, y^{\mathrm{new}}_m)]_+$, where $[\cdot]_+$ denotes $\max(0, \cdot)$.
While frameworks like Optuna~\citep{akiba2019optuna} and BoTorch~\citep{balandat2020botorch} employ HBDA to accelerate the BO sampling routine, the exact algorithm scales poorly.
The number of hyperrectangles $K$ grows as $O(N^{\floor{\frac{M+1}{2}}})$~\citep{lacour2017box}, rendering HBDA intractable for large $N$ and $M$.

\cite{couckuyt2012towards} addressed this bottleneck by an approximation heuristic (Algorithm~\ref{alg:approx-hbda}) that prunes negligible hyperrectangles.
Let $H^{\mathrm{all}}$ denote the total HV of the non-dominated space $\mathcal{N}$, and $H_k$ the HV of a specific box $B_k$.
The algorithm discards any $B_k$ where $H_k \leq \alpha H^{\mathrm{all}}$ (see Line~\ref{line:pruning-conds}), using a control parameter $\alpha \in (0, 1)$.
Note that setting $\alpha = 0$ recovers the exact HBDA.
This heuristic guarantees that the number of hyperrectangles $K$ remains below $\frac{2}{\alpha}$.
Notably, BoTorch currently utilizes defaults of $\alpha = 10^{-2}$ for $M \geq 6$ and $\alpha = 10^{-3}$ otherwise~\footnote{\url{https://github.com/meta-pytorch/botorch/blob/v0.16.1/botorch/acquisition/multi_objective/utils.py\#L63}}.
A visual representation of this process is provided in Figure~\ref{fig:box-decomp-algo}.
The time complexities for the preprocessing (Algorithm~\ref{alg:approx-hbda}) and the HV improvement calculation are $O(\frac{M^2N}{\alpha}\log N)$ and $O(\frac{M}{\alpha})$, respectively.
Note that the former is mostly $O(\frac{MN}{\alpha})$ in practice.
See Appendix~\ref{sec:proof-of-time-complexity} for the proof.

\begin{algorithm}[tb]
  \caption{Approximation of Hypervolume Box-Decomposition Algorithm}
  \label{alg:approx-hbda}
  \begin{algorithmic}[1]
    \Statex{$\mathcal{P}$ (the Pareto set), $\alpha \in )(0, 1)$ (relative volume tolerance).}
    \LineComment{Recall $f_m^{(0)} = f_m^{(1)} - 1, f_m^{(N+1)} = f_m^{(N)} + 1$.}
    \State{Get $\{\{f^{(n)}_m\}_{m=1}^M\}_{n=0}^{N+1}$ by sorting $\mathcal{P}$ with respect to each objective.}
    \State{$H^{\mathrm{tol}} \leftarrow \alpha H^{\mathrm{all}} \coloneqq \alpha \prod_{m=1}^{M} (f_{m}^{N+1} - f_{m}^{0})$, \texttt{stack} $\leftarrow \{\{(0, N+1)\}_{m=1}^M\}, B \leftarrow \emptyset$.}
    \While{\texttt{stack} is not empty}
    \State{Pop $\{(i_m, j_m)\}_{m=1}^M$ from \texttt{stack}, and define $\boldsymbol{l} = [f_1^{(i_1)}, \dots, f_M^{(i_M)}]$ and $\boldsymbol{u} = [f_1^{(j_1)}, \dots, f_M^{(j_M)}]$.}
    \If{$\boldsymbol{u} \nprec \mathcal{P}$}
    \State{Append $\prod_{m=1}^M [l_m, u_m]$ to $B$.}
    \label{line:acceptance}
    \State{\textbf{continue}}
    \EndIf
    \State{$m^\star \in \argmax_{m \in [M]} j_m - i_m$, $H \leftarrow \prod_{m=1}^{M} (u_m - l_m)$.}
    \If{$\mathcal{P} \prec \boldsymbol{l}$ or $j_{m^\star} - i_{m^\star} \leq 1$ or $H \leq H^{\mathrm{tol}}$}
    \label{line:pruning-conds}
    \LineComment{Strictly speaking, $\mathcal{P} \npreceq \boldsymbol{l}$ considering equalities.}
    \State{\textbf{continue}}
    \EndIf
    \State{Copy $\{(i_m, j_m)\}_{m=1}^M$ to $I^{\mathrm{low}}$ and $I^{\mathrm{high}}$, and define $i_{\mathrm{mid}} = \floor{\frac{i_{m^\star} + j_{m^\star}}{2}}$.}
    \State{Update $i_{m^\star}$ in $I^{\mathrm{high}}$ and $j_{m^\star}$ in $I^{\mathrm{low}}$ to $i_{\mathrm{mid}}$, and add $I^{\mathrm{low}}$ and $I^{\mathrm{high}}$ to \texttt{stack}.}
    \EndWhile
    \State{\textbf{return} $B$}
  \end{algorithmic}
\end{algorithm}

\begin{figure}
  \centering
  \includegraphics[width=0.98\textwidth]{figs/box-decomp-algo.pdf}
  \caption{
    Step-by-step visualization of Algorithm~\ref{alg:approx-hbda}.
    $\mathcal{P} = \{[2, 8], [6, 4], [8, 2]\}$, $\rv = [10, 10]$, and $\alpha = 0.1$ are used for this example.
    Red represents the rectangle to split in the current iteration, while blue means these rectangles are in the stack.
    Green is accepted in Line~\ref{line:acceptance} owing to the dominance $\boldsymbol{u} \nprec \mathcal{P}$, and gray is discarded.
    The black rectangles in the final result are the missed rectangles compared to the exact HBDAs.
    The summation of these rectangles' HVs is the worst-case error in the HV improvement approximation.
  }
  \label{fig:box-decomp-algo}
\end{figure}

\section{Related Work \& Conclusion}

This paper provided a rigorous algorithmic description of the approximated HBDA.
Although approximation is essential for making HV-based BO viable in many-objective settings, the literature remains dominated by exact computation algorithms, as surveyed by \cite{guerreiro2021hypervolume}.
While several stochastic Monte Carlo-based approximations exist \citep{bader2011hype,bringmann2012approximating}, \cite{couckuyt2012towards} represent one of the few deterministic alternatives.
Crucially, however, we observe that neither approach guarantees a non-zero HV improvement.
This ``vanishing gradient'' problem highlights the necessity of future work focused on increasing the robustness of these approximations for optimizer feedback.

\bibliographystyle{apalike}  % <==== Change the style file path
\bibliography{ref}

\newpage

\appendix

\section{Proofs of Time Complexities}
\label{sec:proof-of-time-complexity}

\begin{proposition}
  The number of hyperrectangles $K$ obtained by Algorithm~\ref{alg:approx-hbda} is bounded by $\frac{2}{\alpha}$.
  \label{proposition:k-is-less-than-2-by-alpha}
\end{proposition}

\begin{proof}
  Owing to the pruning condition $H \leq \alpha H^{\mathrm{all}}$, the HV of the parent hyperrectangle of an accepted hyperrectangle is always greater than $\alpha H^{\mathrm{all}}$.
  Since each accepted hyperrectangle has its own parents, the summation of HVs for each parent (allowing duplications) is greater than $K \alpha H^{\mathrm{all}}$.
  The upper bound of the summation is obtained when each parent appears twice in its calculation, and the parent set covers $H^{\mathrm{all}}$.
  Note that each parent spawns two children, so they cannot appear more than twice in the calculation.
  Therefore, $K \alpha H^{\mathrm{all}} \leq 2H^{\mathrm{all}} \Longrightarrow K \leq \frac{2}{\alpha}$ holds. 
  This completes the proof.
\end{proof}

\begin{proposition}
  The time complexity of the HV improvement calculation using the result by Algorithm~\ref{alg:approx-hbda} is $O(\frac{M}{\alpha})$.
\end{proposition}

\begin{proof}
  Owing to Proposition~\ref{proposition:k-is-less-than-2-by-alpha}, $K \leq \frac{2}{\alpha}$, while the HV improvement formulation requires $O(MK)$.
  Putting them together, the time complexity is $O(\frac{M}{\alpha})$.
  This completes the proof.
\end{proof}

\begin{proposition}
  The time complexity of Algorithm~\ref{alg:approx-hbda} is $O(\frac{M^2N}{\alpha}\log N)$.
\end{proposition}

\begin{proof}
  The overall time complexity is the summation of the sorting cost and the recursive box decomposition cost.
  The sorting step requires $O(MN \log N)$.
  We now focus on the recursive part.
  As the dominance checks $\boldsymbol{u} \nprec \mathcal{P}$ and $\mathcal{P} \prec \boldsymbol{l}$ constitute the computational bottleneck with a cost of $O(MN)$ per check, the recursive time complexity is $O(MNC)$, where $C$ denotes the maximum iteration count.
  In the worst-case scenario, $C$ is bounded by the product of the maximum recursion depth and the number of accepted hyperrectangles.
  Based on Proposition~\ref{proposition:k-is-less-than-2-by-alpha}, the number of hyperrectangles is bounded by $O(\frac{1}{\alpha})$.
  Due to the binary partitioning of sorted indices, the recursion depth is at most $O(\log N)$ for each objective, resulting in a maximum total depth of $O(M \log N)$.
  Consequently, the overall time complexity is $O(MN \log N + \frac{M^2N}{\alpha} \log N) = O(\frac{M^2N}{\alpha} \log N)$.
  This completes the proof.
\end{proof}

Notice that if the recursion is balanced--which is common in practice--the maximum iteration count is proportional to the number of hyperrectangles, i.e., $C \approx 2 \cdot O(\frac{1}{\alpha})$.
In this case, the time complexity simplifies to $O(\frac{MN}{\alpha} + MN \log N) \approx O(\frac{MN}{\alpha})$, assuming the decomposition cost dominates the sorting cost.

% Automatically include or exclude the appendix information based on paper-status.tex
% \input{settings/appendix-information.tex}

\end{document}